\pdfoutput=1

\documentclass[11pt]{article}
\usepackage[final]{acl}
\usepackage{amssymb}
\usepackage{amsmath}
\usepackage{multirow}
\usepackage{booktabs}
\usepackage{adjustbox}
\usepackage{caption}
\usepackage{times}
\usepackage{latexsym}

\usepackage[T1]{fontenc}

\usepackage{enumitem}
\usepackage[utf8]{inputenc}
\usepackage{microtype}
\usepackage{inconsolata}
\usepackage{graphicx}
\usepackage{comment}

\title{Semantic Consistency-Based Uncertainty Quantification for \\Factuality in Radiology Report Generation}
\author{
Chenyu Wang$^1$, Weichao Zhou$^2$, Shantanu Ghosh$^1$, Kayhan Batmanghelich$^1$, Wenchao Li$^1$ \\
$^1$Boston University, $^2$Massachusetts Institute of Technology \\
\texttt{\{chyuwang, shawn24, batman, wenchao\}@bu.edu, zhouw534@mit.edu} \\
}

\excludecomment{zwccomment}

\begin{document}
\maketitle

\begin{abstract}
\label{sec:abstract}
Radiology report generation (RRG) has shown great potential in assisting radiologists by automating the labor-intensive task of report writing. While recent advancements have improved the quality and coherence of generated reports, ensuring their factual correctness remains a critical challenge. Although generative medical Vision Large Language Models (VLLMs) have been proposed to address this issue, these models are prone to hallucinations and can produce inaccurate diagnostic information. To address these concerns, we introduce a novel Semantic Consistency-Based Uncertainty Quantification framework that provides both report-level and sentence-level uncertainties. Unlike existing approaches, our method does not require modifications to the underlying model or access to its inner state, such as output token logits, thus serving as a plug-and-play module that can be seamlessly integrated with state-of-the-art models. Extensive experiments demonstrate the efficacy of our method in detecting hallucinations and enhancing the factual accuracy of automatically generated radiology reports. By abstaining from high-uncertainty reports, our approach improves factuality scores by $10$\%, achieved by rejecting $20$\% of reports using the \texttt{Radialog} model on the MIMIC-CXR dataset. Furthermore, sentence-level uncertainty flags the lowest-precision sentence in each report with an $82.9$\% success rate. Our implementation is open-source and available at \href{https://github.com/BU-DEPEND-Lab/SCUQ-RRG}{https://github.com/BU-DEPEND-Lab/SCUQ-RRG}.


\end{abstract}

\section{Introduction}
\label{sec:intro}
RRG is gaining importance as healthcare demands grow, placing substantial pressure on radiologists to interpret medical images swiftly and accurately. Automating the report-writing process holds the potential to alleviate this burden, improving both efficiency and diagnostic precision. Vision Large Language Models (VLLMs) have introduced new possibilities in this area by generating detailed and coherent reports from medical images, providing significant assistance to radiologists~\citep{thawkar2023xraygpt,pellegrini2023radialog}. However, despite these advancements, challenges persist—particularly in ensuring the factual accuracy of these generated reports. A notable issue with VLLMs is their tendency to produce ``hallucinations'', or information that is ungrounded in the visual data or inconsistent with established medical knowledge. For example, a model might incorrectly generate findings such as a diagnosis of pneumonia when none is present ~\cite{hartsock2024vision}, or fabricate prior medical history that does not exist ~\cite{ramesh2022improving,tanida2023interactive,hyland2023maira}. Such hallucinations can lead to inaccurate or misleading diagnostic information, posing significant risks in clinical settings. 

Recent studies have explored various methods to address hallucinations in radiology report generation. ~\citet{ramesh2022improving} utilize a GPT-3-based rewriting technique and a BioBERT-based token classification system to remove references to non-existent prior reports. ~\citet{banerjee2024direct} employ Direct Preference Optimization (DPO) to suppress hallucinated prior exams, significantly reducing such errors while maintaining clinical accuracy. However, these methods remain limited in scope, focusing solely on specific hallucinations, namely hallucinated prior exams, and do not enhance the broader factual accuracy of diverse clinical entities critical for dependable diagnostics.~\citet{bannur2023learning,bannur2024maira} tackle hallucinations by integrating current and prior images with detailed report sections, thereby improving the alignment between generated text and visual data to reduce errors and enhance report consistency. These approaches offer a more comprehensive solution than methods targeting specific hallucination types. However, they rely on specialized architectures and additional training resources, limiting their flexibility and applicability across diverse models. 

Addressing the limitations of prior approaches, our framework provides a plug-and-play solution that mitigates hallucinations through uncertainty quantification (UQ), requiring no architectural modifications or additional training. Broadly compatible with diverse VLLM-based RRG models, it emphasizes semantic consistency between generated content and sampled counterparts. Specifically, our UQ framework assesses the consistency of clinical entities within generated reports, assigning high uncertainty to content with low factual precision. We measure this consistency by comparing clinical facts from the original report with those in multiple sampled reports generated from the same query, relying solely on API-level access to broaden applicability. By abstaining from high-uncertainty reports, we enhance the clinical efficacy of generated outputs. Additionally, by flagging high-uncertainty sentences, we guide radiologists to areas needing further validation, reducing their workload and supporting more accurate interventions.
In summary, our contributions are as follows:
\begin{enumerate}[leftmargin=18pt, noitemsep,topsep=0pt,parsep=0pt,partopsep=0pt,label={(\arabic*)}]
\item We propose a plug-and-play UQ framework that does not require modifications to the internal mechanisms of the model and can be easily integrated with state-of-the-art RRG systems.

\item We propose two domain-specific uncertainty quantification methods for report- and sentence-level analysis to identify clinical content with low semantic consistency, improving the factual accuracy of the generated report. 

\item Our framework improves factuality by abstaining from high-uncertainty reports, achieving a 10\% boost in factuality scores by rejecting 20\% of reports using the \texttt{Radialog} model. Additionally, it flags sentences with the highest uncertainty, accurately identifying those with the lowest factual precision at 82.9\%.

\item We evaluate our framework’s effectiveness in detecting non-existent prior exams and investigate its alignment with factuality across various pathology subgroups.
\end{enumerate}

\begin{figure*}[!t]
  \centering
  \vspace{-1em} 
  \includegraphics[width=1.0\linewidth]{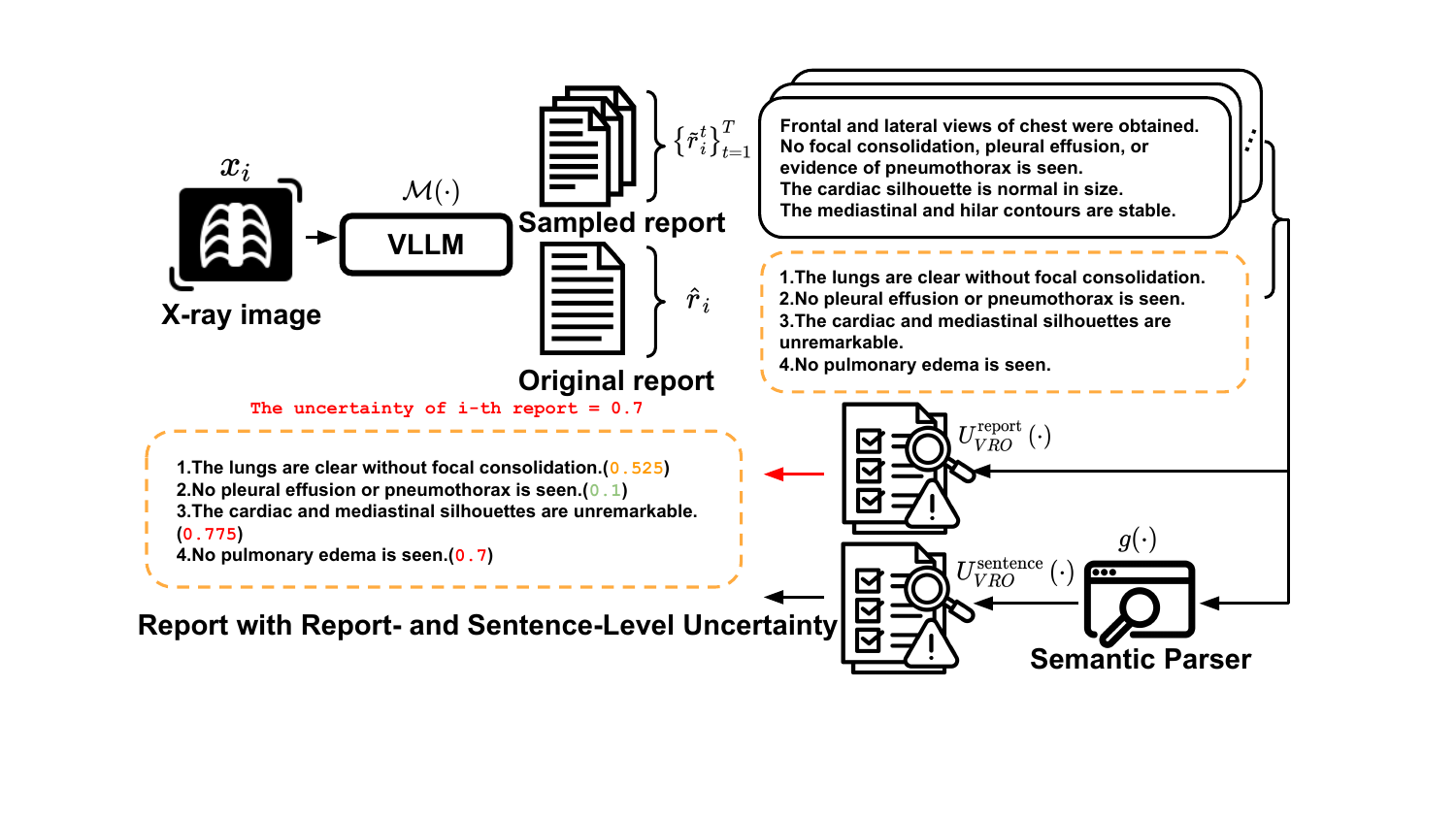} 
  \caption{
  Pipeline of proposed Uncertainty Quantification Framework. Given an X-ray image $x_{i}$, the LLM generates an original report $\hat{r}_i$ and sampled reports $\{\tilde{r}_i^t\}_{t=1}^T$. These reports are first processed by a semantic parser $g$, which extracts entity-label pairs for each sentence in $\hat{r}_i$.  The uncertainty quantification module evaluates semantic consistency at both the report and sentence levels, providing a comprehensive, layered view of uncertainty for the generated report.}
  \label{fig:twocolplot}
\end{figure*}

\section{Preliminaries}
\label{sec:preliminary}
\subsection{RRG with VLLMs}
\label{subsec:background}
In RRG using VLLMs, the input is a medical image $x \in \mathbb{R}^{D}$, where $D$ is the dimension of the image, and the output is a generated report $\hat{r} \in \mathcal{V}^\star$, with $\mathcal{V}^\star$ represents the space of token sequences. To produce this report, the model processes the image through a series of transformations across three main components.

First, the \textbf{image encoder} extracts the visual tokens $f_x = \text{Enc}(x)$ from the image $x$. Next, an \textbf{Alignment Module} generates an embedding $z = g(f_x)$ to map these visual tokens to a text-compatible space. This alignment allows the visual data to be effectively interpreted by the language model (LM). Finally, the aligned embeddings $z$ are passed to a \textbf{Large Language Model (LLM)}, $\mathcal{M}$, which generates the radiology report as a sequence of tokens $\hat{r} \in \mathcal{V}^\star$.


The quality of the generated report $\hat{r}$ is evaluated against a reference report $r$ using a correctness function $A(\hat{r}, r)$, which measures lexical or semantic similarity, assessing how well the generated report aligns with the ground truth.

\subsection{Rank Calibration}
\label{subsec:rankc}
Rank Calibration \cite{huang2024uncertainty} is designed to evaluate the alignment between the uncertainty levels of an LM's predictions and their actual (in)correctness. 
An uncertainty measure is considered rank-calibrated if predictions with higher uncertainty are more likely to be incorrect.
Given $N$ predictions by the LM, each associated with an uncertainty score $u_i$ for $i=1,2,\ldots, N$, these scores are evenly partitioned into $B$ intervals $\{\mathcal{I}_b\}_{b=1}^B$, such that each interval contains approximately $N/B$ scores. 
Using a regression function $reg$ to map the uncertainty score $u$ from any interval $\mathcal{I}_b$ to the accuracy of predictions in that interval, the Empirical Rank-Calibration Error (RCE) assesses the alignment between uncertainty and accuracy as below.
Lower Empirical RCE values indicate better calibration.

\begingroup
\fontsize{9.2}{10}\selectfont 
\begin{align} 
\label{eq:rce} 
RCE &= \frac{1}{B} \sum^B_{b=1}\Bigg|\frac{\sum^B_{\substack{b'=1\\ b'\neq b}} \mathbf{1}\left[\underset{u'\in \mathcal{I}_{b'}}{\sum} reg(u') \geq \underset{u\in \mathcal{I}_{b}}{\sum} reg(u)\right]}{B-1}\notag \\
    & \qquad - \quad\frac{\sum^B_{\substack{b'=1\\ b'\neq b}}  \mathbf{1}\left[\underset{u'\in \mathcal{I}_{b'}}{\sum} u' \leq \underset{u\in \mathcal{I}_{b}}{\sum} u\right]}{B-1}\Bigg| 
\end{align}
\endgroup

\subsection{VRO (Variation Ratio for Original Prediction)
}
The VRO metric~\cite{huang2023look} measures uncertainty by comparing the model’s original prediction with the predictions generated from multiple stochastic inferences. It is calculated as:
\begin{equation}
\label{eq:vro}
VRO = 1 - \frac{1}{T} \sum_{i=1}^{T} \left(1 - \text{dist}(p_i, p_{LM})\right)
\end{equation}
where $T$ is the number of inferences, $p_i$ is the prediction from the $i$-th inference, and $p_{LM}$ is the original prediction from the model. The function $\text{dist}(\cdot)$ measures the distance between two predictions.
Lower VRO values indicate greater consistency between the original and sampled predictions, signifying lower uncertainty in the model’s output.
\subsection{Radgraph}
\label{subsec:radg}
RadGraph~\cite{jain2021radgraph} structures chest X-ray reports by extracting clinical entities and their relationships as multiple triplets. Entities include Anatomy (ANAT-DP) and three types of Observation: Definitely Present (OBS-DP), Uncertain (OBSU), and Definitely Absent (OBS-DA). Anatomy refers to body parts like ``lung,'' while Observations describe features or diagnoses, such as ``effusion'' or ``increased.'' Relations between entities are categorized as \texttt{Suggestive Of}, \texttt{Located At}, or \texttt{Modify}, indicating how observations are inferred, located, or modified. To perform this extraction, \texttt{PubMedBERT}~\cite{gu2021domain}, a pre-trained biomedical language model, was fine-tuned on the RadGraph dataset. It processes radiology report text to automatically label entities and relations, enabling structured analysis of the clinical content.

\subsection{Natural Language Inference
based Uncertainty Quantification}
\label{subsec:nli}
A Natural Language Inference (NLI) model takes a pair of sentences (a premise and a hypothesis) as input and outputs logits for the labels—entailment, contradiction, or neutral—indicating the likelihood of each relationship. \citet{kuhn2023semantic,lin2023generating} leverage these pairwise similarity scores to assess the consistency between response pairs and use them for subsequent uncertainty estimation.
\citet{zhang2024luq} use an off-the-shelf \texttt{DeBERTa-v3-large} model \citep{he2021debertav3} to compute NLI-based uncertainty for each sentence $s_j$ in a response. They calculate the probability of "entailment" by normalizing the entailment logit $l_e$ over the sum of entailment and contradiction logits:
\begin{equation}
P(\text{entail} \mid s_j, r') = \frac{\exp(l_e)}{\exp(l_e) + \exp(l_c)}
\end{equation}
In this way, sentence-report similarity can be calculated to enable UQ.

\section{Method}
\label{sec:method}
 \label{subsec:problem}
In RRG, given an LLM $\mathcal{M}$, we use $\tilde{r}_{i}^{t} = \mathcal{M}_{t}(x_{i})$ to denote $t$-th sampled report given a Chest X-ray image $x_{i}$ in contrast to $\hat{r}_{i} = \mathcal{M}(x_{i})$ as original report. An uncertainty measure is defined as $U^{\mathcal{M}}: \mathcal{V}^\star \times 2^{\mathcal{V}^\star} \to \mathbb{R}$, takes the original report and a set of sampled reports as input and outputs a real value representing the uncertainty. Our core principle is that higher uncertainty should correspond to lower quality in generated outputs. 
Therefore, we follow the approach as described in Section \ref{subsec:rankc} to evaluate the alignment between $U^{\mathcal{M}}$ and the prediction correctness indicated by a clinical metric $F$ via Equation~\ref{eq:rce} by defining the regression function as $reg(u) = \mathbb{E}[F | U^{\mathcal{M}} = u]$.  
However, the long-form nature of RRG poses the following challenges in designing the uncertainty measurement $U$\footnote{Will omit $\mathcal{M}$ when the choice of the LM is clear.}.

\begin{enumerate}[leftmargin=18pt,itemindent=0pt,noitemsep,topsep=0pt,parsep=0pt,partopsep=0pt,label={\bf (\alph*)}]
\item {\bf High Similarity Across Responses:} long texts often yield high similarity across response pairs~\citep{zhang2024luq}, limiting UQ methods based on response-level similarity~\cite{kuhn2023semantic,lin2023generating}. Applying similarity at the component level requires extra effort to align corresponding parts, as sampled responses may reorder or omit claims.
\item {\bf Lack of Domain-Specific NLI Models:} ~\citet{zhang2024luq} propose using NLI models for nuanced similarity assessments; however, RRG lacks a specialized NLI model. General NLI models often struggle with the domain's subtle distinctions, causing error propagation. While \citet{bannur2024maira} leverage the in-context learning abilities of GPT-4 and Llama3-70B for entailment verification—potentially making them viable as UQ methods in RRG—these models are impractical for real-time UQ due to high computational demands. See further discussion in Appendix~\ref{sec:appendixc_subgroup}.

\item {\bf Limitations of Self-Evaluation-Based UQ:} Self-evaluation UQ methods~\citep{kadavath2022language, lin2023generating} attempt to verbalize confidence through handcrafted prompts, enabling models to express uncertainty in natural language. However, this approach is currently unavailable for VLLM-based RRG models\citep{gui2024conformal}, with failure cases demonstrated in the Appendix~\ref{sec:appendixc_subgroup}.
\end{enumerate}
\begin{zwccomment} 
\subsection{Sampling}
\label{subsec:sampling}
The variance observed among multiple responses to a given image provides valuable insight into model confidence. Following previous work ~\cite{zhang2024luq, kuhn2023semantic, xiong2023can}, we repeatedly query the model with the same input using multinomial sampling to generate multiple outputs from the auto-regressive model’s distribution. This enables the evaluation of semantic similarity and uncertainty across predictions.
\subsection{Semantic Equivalence}
\label{subsec:semantic}
\end{zwccomment}

To overcome these challenges, we propose to quantify uncertainty by evaluating semantic similarity between paired reports with clinical metric $F$. 
By focusing on semantic consistency, our method more effectively captures semantic equivalence, leading to improved uncertainty estimation. 
In addition, we apply VRO~\citep{huang2023look}, which calculates the similarity between the original and sampled predictions, to enhance computational efficiency.
In contrast to previous methods~\citep{zhang2024luq, kuhn2023semantic} that require $O(n^2)$ calls to NLI models for pairwise comparisons, our approach reduces this complexity to $O(n)$ calls for consistency measurement while maintaining good performance in UQ with different granularity.
In Section \ref{subsec:reportu} we will provide details on report-level UQ, while Section \ref{subsec:setenceu} details sentence-level UQ.

\subsection{Report-Level Uncertainty Quantification}
\label{subsec:reportu}
Our report-level uncertainty quantification leverages the approach in Equation \ref{eq:vro}
, where we use a factual metric in RRG as the distance function. In this setup, $\hat{r}_{i}$ is treated as the original prediction, and $\tilde{r}^{t}_i$ represents the $t$-th sampled prediction. The uncertainty is computed as:
\begin{equation}
U^{\text{report}}_{\text{VRO}}(\hat{r}_{i},\{\tilde{r}_i^t\}_{t=1}^T)=\frac{1}{T}\sum_{t=1}^T\left(1-F\left(\hat{r}_{i},\tilde{r}^{t}_i \right)\right)
\end{equation}
We leverage GREEN ~\citep{ostmeier2024green}, a state-of-the-art evaluation metric that aligns with radiologist preferences, to implement $F$. GREEN calculates factual alignment by comparing findings and error counts between reports. Here, the original report serves as the prediction, and the sampled reports are references, effectively capturing semantic equivalence between the original generated and sampled reports. Further details on GREEN are in the Appendix \ref{sec:appendixb_subgroup}.

\subsection{Sentence-Level Uncertainty Quantification}
\label{subsec:setenceu}
While report-level uncertainty quantification is useful, it can obscure variations in certainty across multiple facts within a report, making sentence-level quantification more appropriate. ~\citet{zhang2024luq} calculates sentence-to-report entailment scores across all sampled reports increases classifier complexity and computational demands, making it inefficient for real-world deployment. To overcome these challenges, we propose a novel method leveraging the RadGraph~\citep{jain2021radgraph} parser. Assume that each report, $\hat{r}_{i} = \{s_{i1},s_{i2},s_{i3}...s_{ik_{i}}\}$, consists of multiple sentences, where $k_{i}$ indicates the number of sentences within the report. We utilize the RadGraph parser, denoted as $g:\mathcal{V}^\star \to \overline{V} $, which map sequence(s) to the set of node-label pairs $\overline{{V}}=\left\{\left(v_k, v_{k_L}\right)\right\}_{k \in[1 . .|V|]}$ where each pair represents an entity and its associated label. An entity $v_k$ is a continuous text span (potentially multi-word) that represents either an Anatomy or an Observation. The label $v_{k_L}$ for each entity $v_k$ indicates one of the four possible entity categories describe in Section~\ref{subsec:radg}.
Using this structured output, we calculate an uncertainty value for each sentence $s_{ij}$, where $s_{ij}$ is the $j$-th sentence in the generated report $\hat{r}_{i}$.

\begingroup
\fontsize{9.2}{10}\selectfont 
\begin{align}
\label{eq:vro_sentence}
U_{\text{VRO}}^{\text{sentence}} \left( s_{ij}, \{\tilde{r}_i^t\}_{t=1}^T \right) 
&= \frac{1}{T} \sum_{t=1}^T \left( 1 - \frac{\left| g\left( s_{ij} \right) \cap g\left( \tilde{r}_i^t \right) \right|}
{\left| g\left( s_{ij} \right) \right|} \right) \nonumber \\
&= \frac{1}{T} \sum_{t=1}^T \left( 1 - \frac{\left| \overline{V}_{ij} \cap \overline{V}_{i}^{t} \right|}
{\left| \overline{V}_{ij} \right|} \right)
\end{align}
\endgroup

where $\overline{V}_{ij}$ represents the set of node-label pairs in the original sentence $s_{i}^{j}$, and $\overline{V}_{i}^{t}$ is the set of node-label pairs in the sampled report $\tilde{r}_i^t$. The term $|\overline{V}_{ij} \cap \overline{V}_{i}^{t}|$ denotes the number of entity-label pairs from the original sentence $\overline{V}_{ij}$ that are also present in $\overline{V}_{i}^{t}$. Additionally, $\left|\overline{V}_{ij}\right|$ represents the total number of node-label pairs in the original sentence $s_{i}^{j}$, which bounds the uncertainty value between 0 and 1.

\section{Experiments}
\label{sec:experiments}
In this section, we aim to answer the following research questions:
\textbf{RQ1.} How well does our proposed UQ align with the factual correctness of the generated reports?
\textbf{RQ2.} Can our UQ  enhance the radiologist's intervention process to improve the factual accuracy of generated reports?
\textbf{RQ3.} Can our UQ detect content referring to non-existent prior information?
\subsection{Setup}
\textbf{Datasets. } Following previous works, we conduct our experiments on MIMIC-XCR~\cite{johnson2019mimic}. We follow the original train-val-test splits. \\
\textbf{Models.} We use \texttt{RaDialog}~\cite{pellegrini2023radialog} as the base model for our uncertainty quantification experiments. This model was selected due to its clean architecture, strong performance on the \texttt{ReXRank}~\cite{rexrank} online benchmark, and ease of reproducibility without data restrictions. To further validate our approach, we also apply our method to \texttt{CheXpertPlus\_mimiccxr}~\cite{chambon2024chexpert}, a top-performing model on the MIMIC-CXR benchmark. For this model, we assume only API access to demonstrate the flexibility, plug-and-play nature, and generalizability of our proposed uncertainty quantification framework to different vision-language model-based radiology report generation systems.

\subsection{RRG Evaluation}
We evaluate our RRG models using four metric categories from the ReXRank benchmark~\cite{rexrank}, supplemented by the state-of-the-art \texttt{GREEN} evaluation. We use \textbf{lexical metrics} such as \texttt{BLEU} and embedding-based \texttt{BERTScore} to assess token-level and semantic similarity. To evaluate pathological and entity-based consistency, we apply \textbf{factuality metrics}, including \texttt{Semb Score} and \texttt{RadGraph Precision, Recall, and F1}. We further assess clinical accuracy with \textbf{RadCliQ}, which combines \texttt{RadGraph F1} and \texttt{BLEU} scores, and the \textbf{GPT-based evaluator} \texttt{GREEN}, which evaluates clinical accuracy by matching findings and counting errors between generated and reference reports. For detailed descriptions of each metric, see Appendix~\ref{sec:appendixd_subgroup}.

\subsection{UQ Evaluation}
In this section, we show how UQ can be evaluated in radiology report generation.
In contrast to typical question-answering tasks where the correctness of a model's prediction is binary, radiology report generation typically involves long-form generation which requires more nuanced evaluation methods for UQ.\\
\textbf{Pearson correlation coefficient.} 
The Pearson correlation coefficient can be used to assess how well uncertainty quantification aligns with the factual correctness of generated reports. By measuring the linear relationship between model uncertainty and report quality, Pearson's coefficient provides insight into whether higher uncertainty corresponds to lower factual accuracy. The Pearson correlation ranges between $-1$ and $1$, where a negative value indicates an inverse relationship. In our setting, we use Pearson's coefficient to evaluate this relationship, with a strong negative correlation suggesting that higher uncertainty signals lower report correctness, aligning with the intended behavior of uncertainty quantification.
 \\
\textbf{Rank calibration error.}
RCE assesses the consistency in ranking, ensuring higher uncertainty corresponds to lower correctness, regardless of a linear relationship. We use the Empirical RCE ~\cite{huang2024uncertainty}, which divides uncertainty values into $B=20$ bins. For each bin, we calculate the expected correctness level and average uncertainty. The Empirical RCE is computed by averaging the rank differences between correctness and uncertainty across these bins as Equation~\ref{eq:rce}, offering a principled approach to measure the alignment between uncertainty and correctness without relying on arbitrary thresholds. \\
\textbf{Abstention.} Abstention allows uncertainty quantification to enhance factual accuracy by rejecting high-uncertainty reports, directing radiologists to focus on certain content. Traditionally, abstention is measured by metrics like AUARC~\cite{huang2024uncertainty}, which evaluates improvement by abstaining from uncertain cases. However, binary metrics like AUARC are inadequate for the nuanced nature of RRG. To address this, we evaluate abstention at the report level, measuring improvements in factuality scores while balancing the trade-off with coverage. This strategy enables targeted intervention by radiologists, focusing their review on areas where factual accuracy may be compromised.\\
\textbf{Uncertainty Precision Alignment.} To evaluate sentence-level UQ in RRG, we calculate a factual precision score for each generated sentence using RadGraph (details in Appendix~\ref{sec:appendixd_subgroup}). We then assess how well high uncertainty scores correspond to sentences with low factual precision within each report. Specifically, we measure the alignment rate between the sentence with the highest uncertainty and the sentence with the lowest factual precision. This alignment metric supports targeted interventions, enabling radiologists to focus on sentences that may require closer review due to potential factual inaccuracies.
\subsection{Hallucination Detection}
 In RRG, references to prior exams are a common form of hallucination~\citep{banerjee2024direct}. In this section, We empirically investigate whether our UQ can effectively detect and flag these hallucinations by assigning them high uncertainty.
Following ~\citet{banerjee2024direct}, we define 43 substrings commonly associated with references to prior exams. For report-level uncertainty, we analyze the changes in the percentage of reports with prior exam references and the average number of hallucinated substrings per report before and after applying different levels of abstention. 

\subsection{UQ Baselines}
We compare our method with the previous uncertainty quantification method. Following ~\citet{kuhn2023semantic}, we use predictive entropy, length-normalised predictive entropy~\cite{malinin2020uncertainty} and lexical similarity~\cite{zhang2024luq,fomicheva2020unsupervised}. We do not compare with methods involving NLI classifiers and self-evaluation-based UQ due to their unavailability in RRG, as discussed in Appendix~\ref{sec:appendixc_subgroup}.
For all experiments, we use the default temperature value $1$ and sample $10$ responses to calculate UQ. 


\section{Results}
\label{sec:results}

\subsection{Alignment with Factuality (RQ1)}

Table~\ref{tab:combined_correlations_stacked} demonstrates that our proposed VRO-GREEN exhibits stronger negative Pearson correlations with factuality metrics across both the \texttt{Radialog Model} and the \texttt{CheXpertPlus\_mimiccxr Model} when compared to baseline UQ methods. In particular, VRO-GREEN achieves high negative correlations on GREEN (-0.5292 for \texttt{Radialog}, -0.4726 for \texttt{CheXpertPlus\_mimiccxr}) and RadCliQ-v0 (-0.4137 for \texttt{Radialog}, -0.3743 for \texttt{CheXpertPlus\_mimiccxr}). This indicates VRO-GREEN's superior capability in aligning uncertainty with factual correctness in radiology report generation. Furthermore, we grouped samples based on the presence of specific pathology findings to examine the correlation between UQ and GREEN for each subgroup for the \texttt{Radialog Model}. Subgroup analysis in Table~\ref{tab:subgroup} reveals variation in correlation strength, particularly in the Pneumothorax subgroup, where the correlation is notably weaker at $-0.08$, likely due to the underrepresentation of Pneumothorax cases (around $1$\% of positive cases in the training set). 

Table~\ref{tab:stacked_rce_no_model_column} further validates VRO-GREEN's alignment effectiveness using Empirical RCE, where it achieves the lowest RCE values on both GREEN (0.015 for \texttt{Radialog}, 0.02 for \texttt{CheXpertPlus\_mimiccxr}) and Negative RadCliQ-v0 (0.02 for \texttt{Radialog}, 0.025 for \texttt{CheXpertPlus\_mimiccxr}). These results confirm VRO-GREEN's superior consistency in aligning uncertainty with factual correctness across multiple metrics.

At the sentence level, the Pearson correlation between sentence-level uncertainty (VRO-RadGraph) and factual precision is strong for both models, with -0.52 for the \texttt{Radialog} model and -0.55 for the \texttt{CheXpertPlus\_mimiccxr} model, indicating effective alignment with factuality at sentence level.

\begin{table*}[ht]
\centering
\scriptsize 
\renewcommand{\arraystretch}{1.2} 
\setlength{\tabcolsep}{4pt} 
\adjustbox{max width=\textwidth}{
\begin{tabular}{lccccccccc}
\toprule
\textbf{Uncertainty Method} & \textbf{BLEU Score} & \textbf{BERTScore} & \textbf{Semb Score} & \textbf{RadGraph Recall} & \textbf{RadGraph Precision} & \textbf{RadGraph Combined} & \textbf{GREEN} & \textbf{-RadCliQ-v0} & \textbf{-RadCliQ-v1} \\
\midrule
\multicolumn{10}{c}{\textbf{Radialog Model}} \\ 
\midrule
\textbf{VRO-GREEN (Ours)}           & -12.15 & \textbf{-40.92} & \textbf{-30.71} & \textbf{-19.95} & \textbf{-34.15} &\textbf{ -27.88 }& \textbf{-52.92} & \textbf{-41.37} & \textbf{-39.46 }\\
\textbf{Predictive Entropy}         & -0.79 & -28.77 & -21.03 & -2.78 & -22.35 & -12.25 & -32.84 & -27.72 & -25.18 \\
\textbf{Normalized Entropy}         & -8.43 & -22.32 & -19.00 & -10.58 & -16.78 & -13.60 & -36.34 & -23.19 & -22.04 \\
\textbf{Lexical Similarity}         & \textbf{-12.42} & -38.13 & -26.80 & -16.67 & -31.46 & -24.95 & -38.75 & -37.21 & -35.62 \\
\midrule
\multicolumn{10}{c}{\textbf{CheXpertPlus Model}} \\ 
\midrule
\textbf{VRO-GREEN (Ours)}           & -9.39 &\textbf{ -34.00} & \textbf{-29.7}2 & -22.70 & \textbf{-30.20 }& \textbf{-27.62} & \textbf{-47.26} & \textbf{-37.43 }& \textbf{-35.51} \\
\textbf{Lexical Similarity}         & -\textbf{15.17 }& -31.48 & -25.25 & \textbf{-23.30} & -27.32 & -26.38 & -34.91 & -32.95 & -32.11 \\
\bottomrule
\end{tabular}
}
\caption{Pearson correlation values (expressed as percentages) for various metrics with different UQ methods across two models: \texttt{Radialog} and \texttt{CheXpertPlus\_mimiccxr}. Stronger negative values indicate better performance. Negative RadCliQ metric are used to align with other metrics in Pearson correlation calculations. For \texttt{CheXpertPlus\_mimiccxr}, we assume API-only access to the model, so only lexical similarity is compared in the table.}
\label{tab:combined_correlations_stacked}
\end{table*}

\begin{table}[h!]
\centering
\resizebox{\columnwidth}{!}{%
\begin{tabular}{lccc}
\hline
\textbf{UQ Method} & \textbf{RCE(GREEN)} & \textbf{RCE(-RadCliQ-v0)} \\
\hline
\multicolumn{3}{c}{\textbf{Radialog Model}} \\
\hline
VRO-GREEN (Ours) & \textbf{0.015} & \textbf{0.02} \\
Predictive Entropy & 0.045 & 0.09 \\
Normalized Entropy & 0.045 & 0.145 \\
Lexical Similarity & 0.045 & 0.04 \\
\hline
\multicolumn{3}{c}{\textbf{CheXpertPlus Model}} \\
\hline
VRO-GREEN (Ours) & \textbf{0.02} & \textbf{0.025} \\
Lexical Similarity & 0.03 & 0.03 \\
\hline
\end{tabular}%
}
\caption{Empirical RCE results for various UQ metrics measured on GREEN and Negative RadCliQ-v0 correctness, with results presented separately for Radialog and CheXpertPlus\_mimiccxr models.}
\label{tab:stacked_rce_no_model_column}
\end{table}

\subsection{Enhancing RRG Intervention(RQ2)}
\begin{figure}[ht]
\includegraphics[width=\columnwidth]{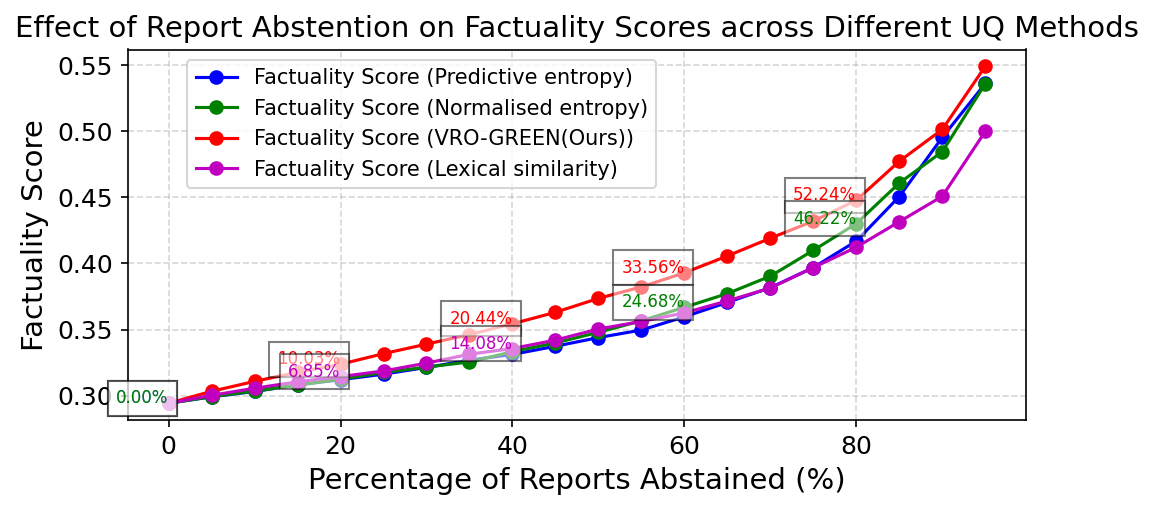}
  \caption{Effect of Report Abstention on Factuality Score across UQ for the \texttt{RaDialog} model. The percentages in boxes represent the improvement(only top-2 visualized) in factuality score after abstention, relative to the initial performance without abstention.}
  \label{fig:abstain_report}
\end{figure}


Figure~\ref{fig:abstain_report} and Figure~\ref{fig:chexpertplus_abstention} illustrate the impact of report-level abstention on factuality scores (GREEN) for the \texttt{Radialog} and \texttt{CheXpertPlus} models, respectively. By excluding the top 20\% most uncertain reports, our UQ method achieves notable factuality improvements: $10$\% for \texttt{Radialog} and $9.2$\% for \texttt{CheXpertPlus}, demonstrating consistent gains across models. These results highlight our method’s effectiveness in enhancing report quality and supporting radiologists in focusing on more reliable reports. 

At the sentence level, uncertainty-precision alignment results reveal that for the \texttt{Radialog} model, the highest-uncertainty sentence aligns with the lowest factual precision at a rate of $82.9$\%, while the lowest-uncertainty sentence aligns with the highest factual precision at only $59.1$\%. For the \texttt{CheXpertPlus} model, these rates are $81.2$\% and $59.6$\%, respectively, closely mirroring the trend observed in \texttt{Radialog}. This discrepancy indicates that while our sentence-level UQ method effectively flags low-precision sentences with high uncertainty, it performs poorly in cases of low-uncertainty sentences, highlighting the presence of confidently hallucinated sentences that our method struggles to capture. This limitation underscores a key challenge in our current approach and suggests an avenue for future work. More details are discussed in Section~\ref{sec:limitation}.
\begin{figure}[t]
\includegraphics[width=\columnwidth]{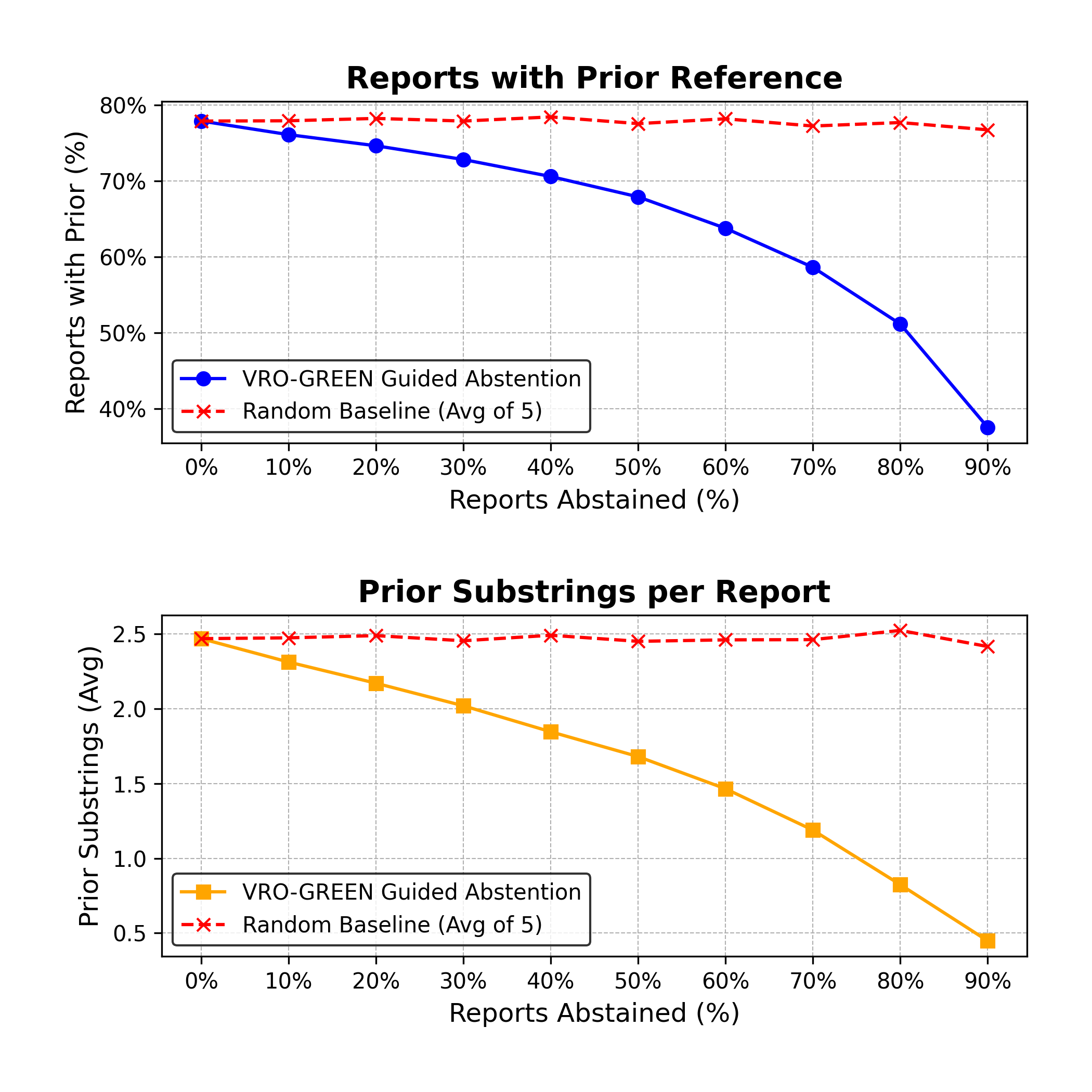}
  \caption{Effect of VRO-GREEN Guided Abstention on Prior References and Substrings for the \texttt{Radialog} model. Solid lines represent VRO-GREEN Guided Abstention, with dashed red lines as the baseline performing random abstention.
  }
  \label{fig:hallucinated-prior-report}
\end{figure}
\subsection{Detection of Hallucinations of Prior Exams (RQ3)}
Figure~\ref{fig:hallucinated-prior-report} and Figure~\ref{fig:chexpert-hallucinated-prior-report} demonstrate the effectiveness of our report-level uncertainty quantification in detecting hallucinations of prior exams for the \texttt{Radialog} and \texttt{CheXpertPlus} models, respectively. Rejecting high-uncertainty reports leads to a clear decrease in the percentage of reports with prior references and the average number of prior-related substrings, significantly improving hallucination detection. In contrast, the random baseline, averaged over 5 seeds, shows no reduction in these metrics.
\begin{figure*}[t]
  \centering
  \vspace{-1.5em}
\includegraphics[width=1.0\linewidth]{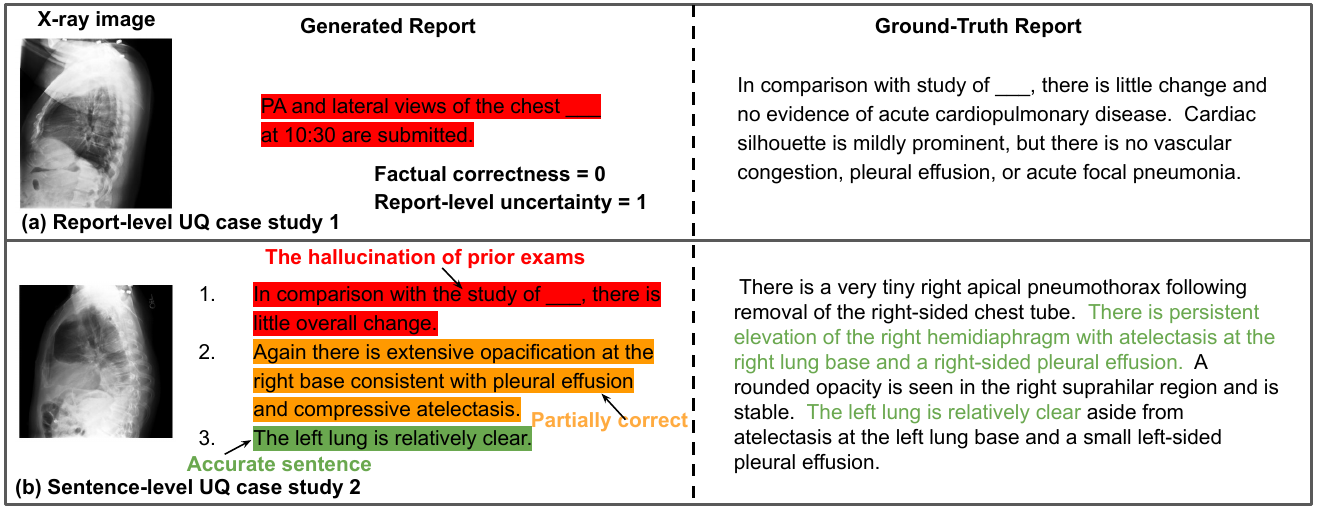} 
  \caption{Two separate analyses of report- and sentence-level UQ in radiology report generation using MIMIC-CXR data. (a) The report-level UQ study assigns an uncertainty score to the entire report. (b) The sentence-level UQ study ranks individual sentences by uncertainty, with red (1.0) indicating high uncertainty, orange (0.75) indicating moderate uncertainty, and green (0.47) indicating low uncertainty. This color-coded ranking helps inform radiologists on which sentences may require closer attention.}
  \label{fig:casestudy}
\end{figure*}
\subsection{Qualitative Analysis.} 

In this section, we analyze the qualitative aspects of our UQ framework for radiology report generation. Specifically, we explore (1) the effect of increasing the number of sampled reports on UQ performance and (2) a case study showcasing the practical utility of our framework in identifying low factual correctness and guiding radiologist interventions.\\
\textbf{Number of sampled reports.} Research on short question-answer tasks and long-form generation has shown that increasing the number of sampled responses can enhance the performance of uncertainty quantification. We extend this investigation to radiology report generation, exploring whether the same holds true in this domain. As illustrated in Figure~\ref{fig:num_samples}, our findings align with previous research, showing that the performance of UQ improves with more samples and converges when using seven samples.\\
\textbf{Case Study.} Figure~\ref{fig:casestudy} illustrates our UQ framework’s ability to identify low factual correctness and assist radiologists in targeted review. In Figure~\ref{fig:casestudy}(a), report-level UQ assigns the highest uncertainty score to a nonsensical report with zero factual correctness. Figure~\ref{fig:casestudy}(b) shows sentence-level UQ, ranking sentences by uncertainty to guide radiologist intervention: the high-uncertainty (red) sentence references a non-existent prior exam, the moderate-uncertainty (orange) sentence is partially correct, and the low-uncertainty (green) sentence is fully accurate.


\section{Related Work}
\label{sec:related}
{\bf Multimodal Foundation Models}, such as VLLMs, augment large language models (LLMs) with visual inputs ~\cite{claude3family,2023GPT4VisionSC}.
These models are typically pre-trained on diverse datasets ~\cite{erhan2010unsupervised,chen2020simple,li2022blip,lin2024vila,alayrac2022flamingo} before applied to specialized tasks, reducing the requirements for domain-specific data. 
VLLMs have been evaluated in medical applications such as medical image interpretation and radiology report generation~\cite{litjens2017survey,esteva2021deep,moor2023foundation,maira2024}, and have demonstrated performance comparable to previous supervised methods~\cite{rajpurkar2017chexnet,qin2018computer}, and in some cases, even rival medical experts ~\cite{tiu2022expert}. 
However, there are challenges that hinder establishing trust in multimodal foundation models in clinical practice ~\cite{truhn2024large,freyer2024regulator,ong2024ethical}. 
These challenges include ensuring the quality and transparency of the training data ~\cite{koccak2022key,celi2022sources,chen2023algorithmic}, effective collaboration between machine learning experts and medical professionals ~\cite{cai2019hello}, and more effective and meaningful evaluation measurements ~\cite{wornow2023shaky}.
Our paper focuses on the factuality of VLLMs' predictions in radiology applications where ensuring accuracy and trustworthiness are critical for clinical decision-making~\cite{bates2021potential}.

\noindent {\bf Hallucination in Foundation Models} \cite{rawte2023survey,ji2023survey} can lead to non-factual predictions, and this issue persists regardless of the model’s size  \cite{lee2023gpt4,jeblick2024chatgpt,xu2024hallucination,zhang2024how}. 
Studies \cite{zheng2023does,lu2023emergent} have shown that a lack of domain-specific knowledge in assigned tasks can cause foundation models to produce hallucinated outputs, a behavior that is often difficult for the models to correct on their own \cite{huang2023large}. 
A large body of research has focused on addressing this issue by filling the knowledge gaps with additional oracle labels \cite{kim2024language,shinn2024reflexion,gou2024critic,banerjee2024direct,bannur2024maira}.
However, as \citet{feng2024dont} points out, the knowledge gaps will always exist because knowledge is continually evolving.
Moreover, in radiology, filling these gaps requires expertise-intensive labeling of data such as medical imaging data \cite{koyyada2023xray,kim2024accurate} and Electronic Health Records (EHR) data \cite{cord23costs}.

\noindent{\bf Uncertainty Quantification (UQ) } has been extensively studied in conventional ML \cite{gupta2006model,shafer2008tutorial,vaicenavicius2019evaluating,tibshirani2019conformal,abdar2021review}, and is receiving increasing attention due to its potential to mitigate hallucinations in foundation models \cite{xiao2021hallucination,fadeeva2024fact}. 
Straightforward methods include querying models about their confidence \cite{xiong2023can,joshi2017triviaqa}, and using Perplexity score \cite{jelinek1977perplexity}. 
Recent research on semantic uncertainty \cite{kuhn2023semantic,zhang2024luq} draw insights from the coherence of model predictions by using an additional NLI \cite{maccartney2009natural} classifier. 
Calibration methods and conformal prediction techniques \cite{liu2024multi,quach2024conformal,gui2024conformal} can offer statistical guarantees on the factuality of the outputs, provided that there is a held-out dataset for extracting necessary information. 
UQ in radiology report generation poses unique challenges to existing methods by involving image inputs and long-form radiologist reports as outputs \cite{koccak2022key,jeblick2024chatgpt,smit2020chexbert}. 
Our method is related to RadGraph \cite{jain2021radgraph}, which structures radiology reports by extracting pre-defined clinical entities and their relations.
Prior works that used RadGraph for UQ involve an additional reinforcement learning step \cite{delbrouck2022improving}.
However, our approach does not involve such step.

\section{Conclusion}
\label{sec:conclusion}
In this paper, we tackle the challenge of hallucinations in RRG through a novel UQ approach. 
Our plug-and-play framework introduces both report-level and sentence-level UQ to detect low-factuality reports and identify non-existent prior hallucinations, supporting more effective radiologist intervention. 
Applied to the MIMIC-CXR dataset, our method achieved a 10\% improvement in factuality by rejecting 20\% of high-uncertainty reports using the \texttt{Radialog} model. Additionally, sentence-level UQ flagged sentences with the lowest factual precision at 82.9\% accuracy, enabling targeted intervention.
Future work will focus on exploring supervised uncertainty measures to improve factuality, particularly addressing cases where the UQ framework assigns low uncertainty to hallucinated predictions generated by VLLMs. Additionally, integrating uncertainty directly into the generation process could guide models toward more factual outputs by conditioning generation on uncertainty thresholds, thus enhancing both the reliability of UQ and overall model trustworthiness.

\section{Acknowledgments}
\label{sec:acknowledgment}
This work was supported in part by the U.S. National Science Foundation under grant CCF-2340776. Additional support was provided by the NIH (Award Number 1R01HL141813-01) and the Pennsylvania Department of Health. We also gratefully acknowledge the computational resources made available by Pittsburgh Super Computing (grant number TGASC170024).

\section*{Limitations}
\label{sec:limitation}
In this section, we outline the limitations of our work and potential areas for improvement.

First, while we demonstrate the effectiveness of our method across different model architectures using the MIMIC-CXR dataset, our evaluation is limited to this dataset. Expanding our experiments to other datasets, such as IU X-Ray~\cite{demner2016preparing} or the recently published CheXpert Plus~\cite{chambon2024chexpert}, could further validate the generalizability of our approach.

Second, due to challenges outlined in Section~\ref{sec:method}, we were only able to compare our method against three relatively simple baselines. As UQ techniques continue to evolve within this domain, the development of domain-specific models, such as tailored NLI models for RRG, could enable a more comprehensive comparison in future work.

Third, while our sentence-level uncertainty quantification effectively aligns high-uncertainty sentences with low factual precision, it struggles to align low-uncertainty sentences with high factual precision, revealing a gap in detecting confidently hallucinated sentences. This limitation suggests the need for enhanced UQ techniques and the potential benefit of incorporating a fact-checking module to improve reliability and distinguish factual inaccuracies.

Finally, our current sentence-level UQ is designed with intervention in mind, focusing solely on the factual precision of generated reports. However, this approach overlooks factual completeness, meaning it does not account for important factual information that may be omitted from the generated report. Future work could address this by designing UQ methods that consider both factual precision and completeness, providing a more balanced evaluation of report quality.

These limitations highlight opportunities for further refinement and experimentation in UQ methodologies for radiology report generation

\appendix
\section{GREEN}
\label{sec:appendixb_subgroup}
Given a generated report as the hypothesis and a ground truth report as the reference, the GREEN evaluation framework assesses clinical accuracy by analyzing both error counts across various clinically significant categories and counts of matched findings. Specifically, GREEN categorizes errors as follows:\\
\textbf{(a)} False report of a finding in the candidate. \\
\textbf{(b)} Missing a finding present in the reference. \\
\textbf{(c)} Misidentification of a finding's anatomic location/position. \\
\textbf{(d)} Misassessment of the severity of a finding. \\
\textbf{(e)} Mentioning a comparison that isn't in the reference. \\
\textbf{(f)} Omitting a comparison detailing a change from a prior study.
The GREEN score is then calculated as:
\begin{equation*}
\text { GREEN }=\frac{\text { \# matched findings }}{\# \text { matched findings }+\sum_{i=(a)}^{(f)} \# \text { error }_{\text i}}
\end{equation*}

\section{Challenges in Applying Other UQ Methods to RRG}
\label{sec:appendixc_subgroup}
\subsection*{NLI-based UQ}
The lack of domain-specific NLI models in RRG makes this approach infeasible. Although \citet{bannur2024maira} leverage in-context learning with large models like GPT-4 and Llama3 for entailment verification in RRG, they are mainly designed to evaluate generated reports against ground-truth reports. Their study reports that RadFact's entailment verification with Llama3-70B requires a single compute node with four A100 GPUs, taking approximately 17 seconds per comparison, while GPT-4, hosted on Microsoft Azure, takes around 27 seconds. Considering the computational requirements discussed in Section~\ref{sec:method}, a single report-level UQ with GPT-4 would require around 675 seconds for five sampled reports, and Llama3's GPU needs make it too costly for UQ applications. All of the above highlights the challenges of applying NLI-based methods for UQ in radiology report generation RRG. Therefore, we call for the development of RRG-tailored NLI models to better support UQ in this domain.
\begin{table}[h!]
\centering
\small
\begin{tabular}{l c}
\hline
    \textbf{Pathology Finding} & \textbf{Pearson Correlation (\%)} \\
    \hline
    No Finding & -68.44 \\
    Enlarged Cardiomediastinum & -42.37 \\
    Cardiomegaly & -35.07 \\
    Lung Opacity & -38.84 \\
    Lung Lesion & -41.47 \\
    Edema & -34.01 \\
    Consolidation & -43.60 \\
    Pneumonia & -38.26 \\
    Atelectasis & -42.91 \\
    Pneumothorax & -8.09 \\
    Pleural Effusion & -27.82 \\
    Pleural Other & -40.04 \\
    Fracture & -46.36 \\
    Support Devices & -42.27 \\
    \hline
\end{tabular}
\caption{Pearson correlation (as percentages) between UQ and GREEN (The overall correlation is -0.52 ) across various subgroups of pathology findings.}
\label{tab:subgroup}
\end{table}

\begin{figure}[ht]
  \includegraphics[width=\columnwidth]{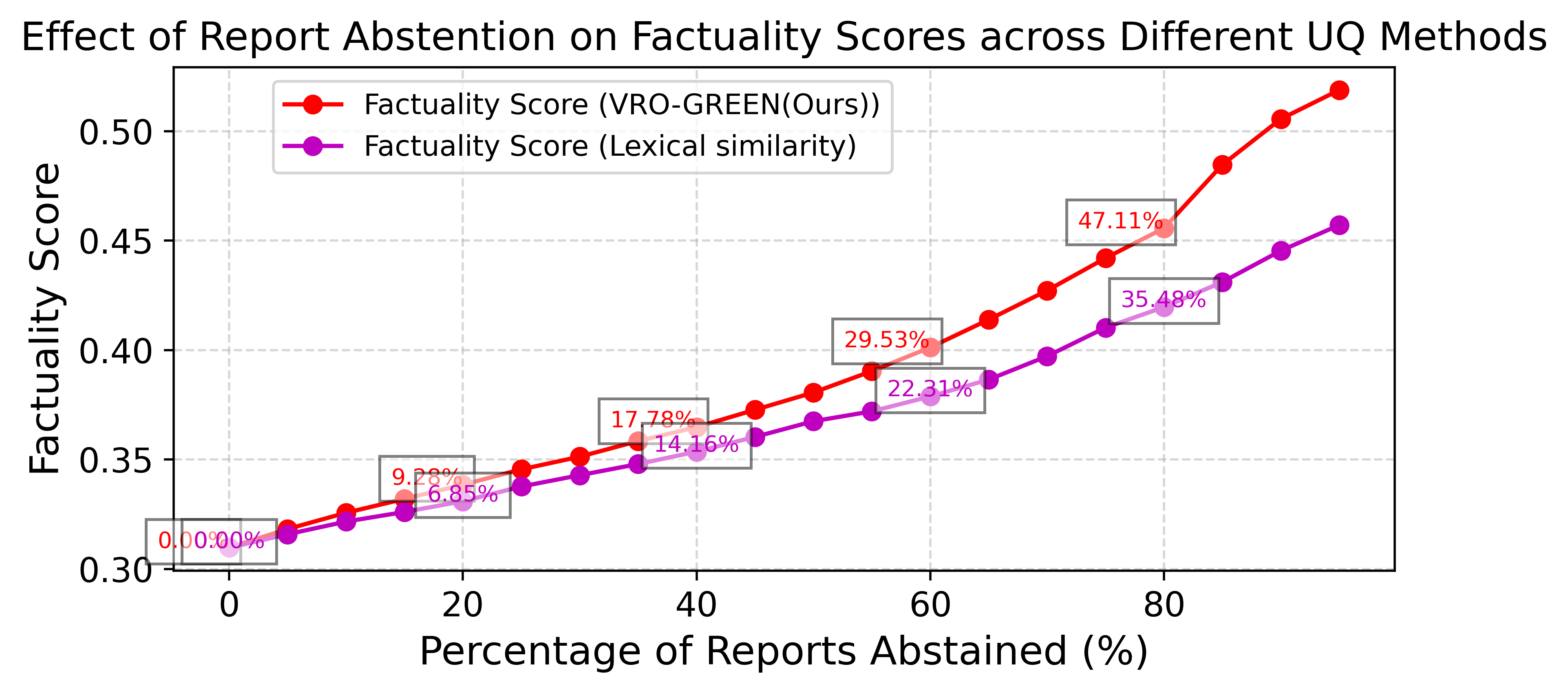}
  \caption{Effect of Report Abstention on Factuality Score across UQ for the \texttt{CheXpertPlus\_mimiccxr} model. The percentages in boxes represent the improvement in factuality score after abstention, relative to the initial performance without abstention. We assume API-only access to the model, so only lexical similarity is compared in the figure.}
  \label{fig:chexpertplus_abstention}
\end{figure}

\begin{figure}[ht]
  \includegraphics[width=\columnwidth]{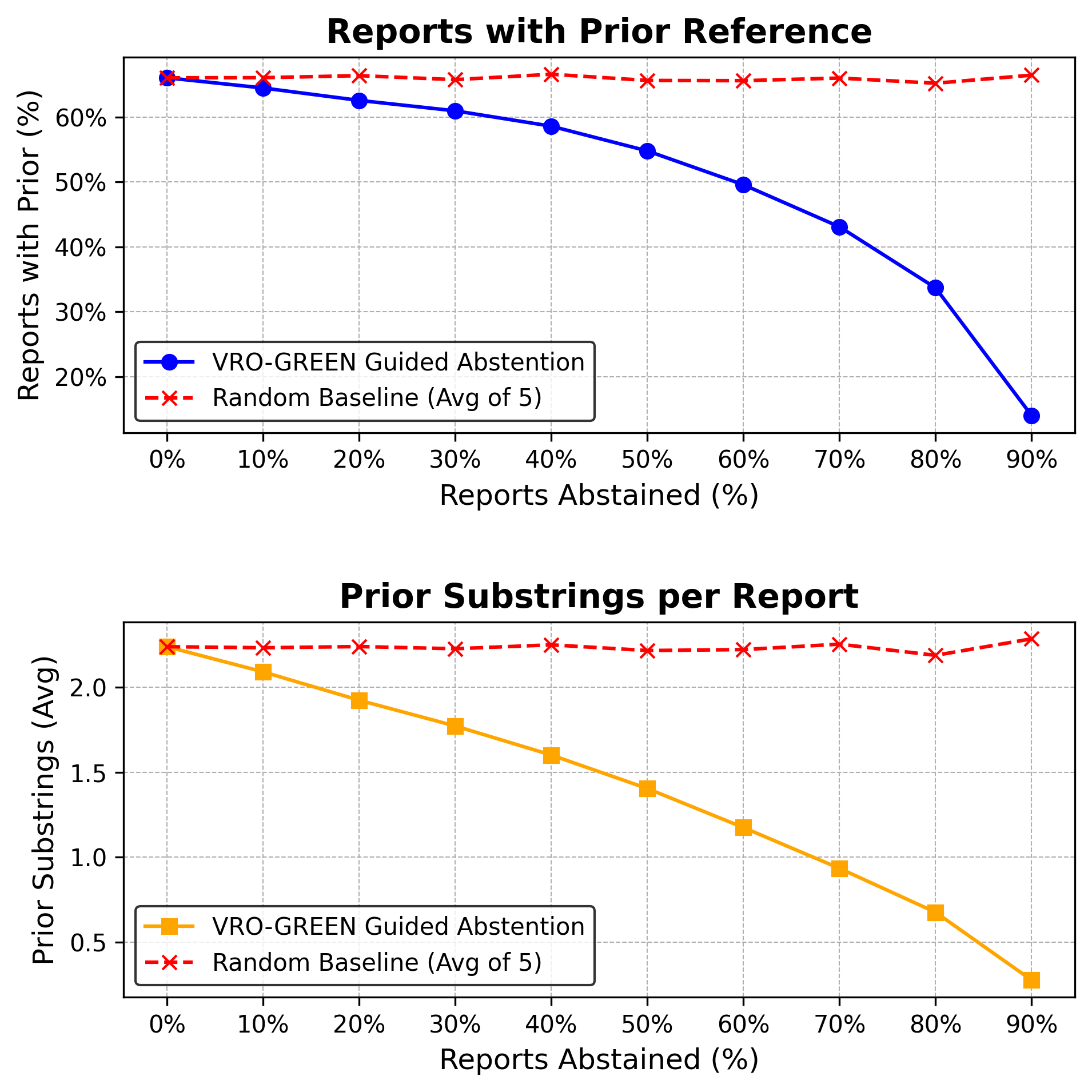}
  \caption{Effect of VRO-GREEN Guided Abstention on Prior References and Substrings for the \texttt{CheXpertPlus\_mimiccxr} model. Solid lines represent VRO-GREEN Guided Abstention, with dashed red lines as a baseline performing random abstention.
  }
  \label{fig:chexpert-hallucinated-prior-report}
\end{figure}

\begin{figure}[ht]
  \includegraphics[width=\columnwidth]{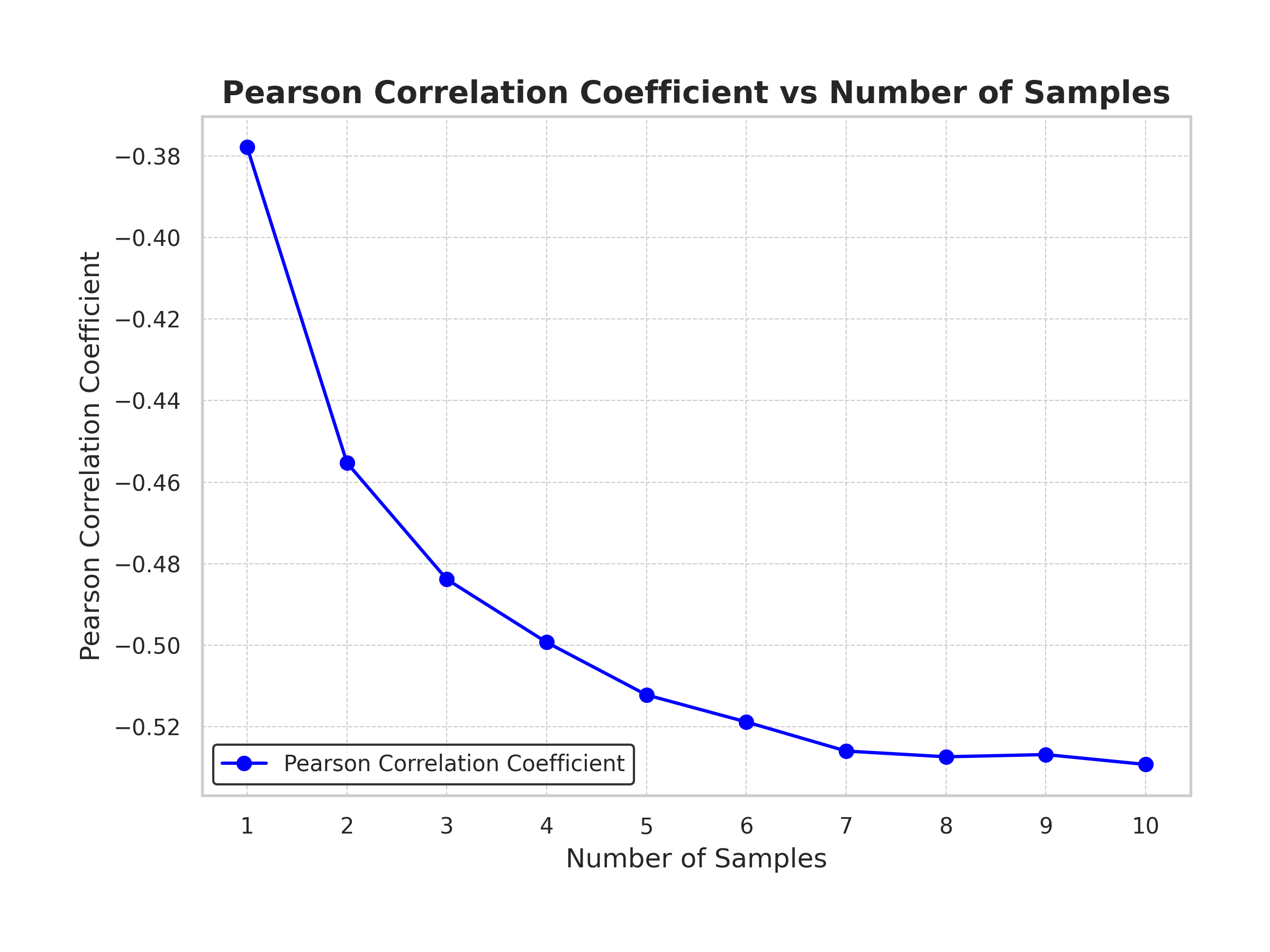}
  \caption{The effect of different number of samples on the Pearson correlation values with VRO-GREEN.}
  \label{fig:num_samples}
\end{figure}

\begin{figure*}[t]
  \centering
\includegraphics[width=1.0\linewidth]{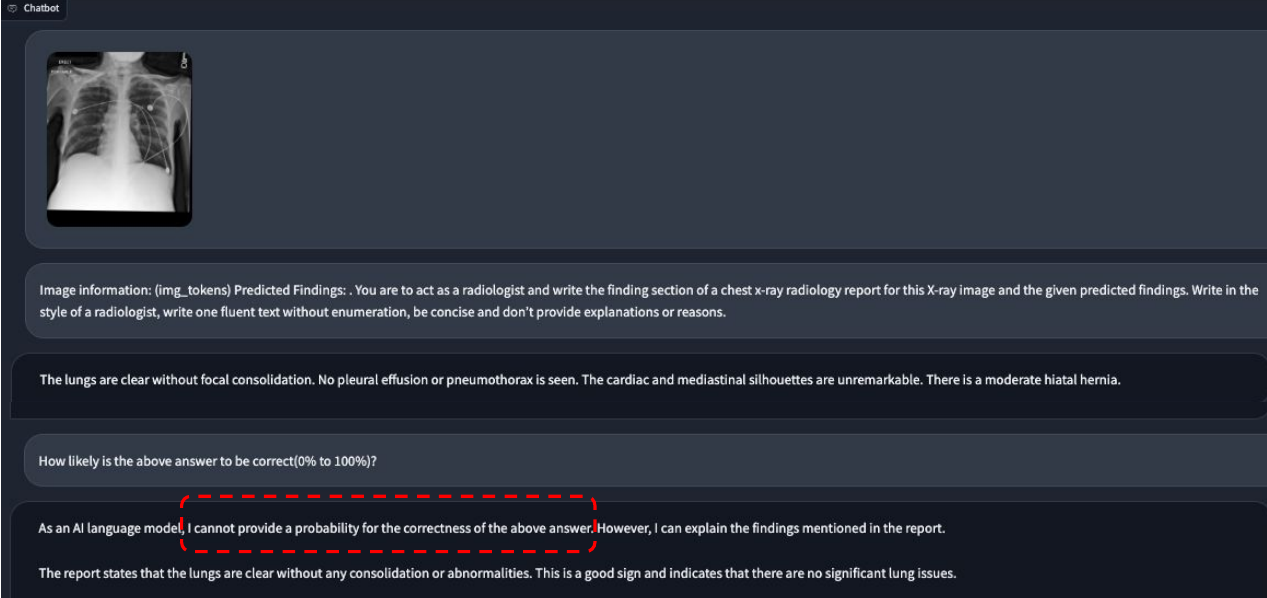} 
  \caption{Example of Failure Cases in Applying Self-Evaluation-Based UQ to the \texttt{RaDialog} Model}
  \label{fig:failure_case}
\end{figure*}
\subsection*{Self-Evaluation-Based UQ}
We demonstrate failure as shown in Figure~\ref{fig:failure_case} cases when applying Self-Evaluation-Based UQ to the \texttt{RaDialog} model in RRG. These limitations likely stem from the model's smaller size, making it less capable of self-probing compared to larger models like GPT-4.

\section{Evaluation}
\label{sec:appendixd_subgroup}

\subsection*{RRG Evaluation}
These metrics are organized into four categories:
\textbf{Lexical Metrics.} We apply traditional Natural Language Processing (NLP) metrics such as \texttt{BLEU} ~\cite{papineni2002bleu} to measure token-level similarity between the generated and ground-truth reports. In addition, we leverage the embedding-based similarity metric \texttt{BertScore} ~\cite{zhang2019bertscore} to capture more nuanced relationships between the texts.\\
\textbf{Factuality Metrics.} To assess factual consistency between the generated and ground-truth reports, we use two key approaches. First, \texttt{Semb Score} is calculated by passing both reports through the \texttt{CheXbert} model ~\cite{smit2020chexbert}, which extracts present/absent/uncertain labels for 14 \texttt{CheXpert} pathological observations ~\cite{irvin2019chexpert}. Cosine similarity between the resulting embeddings is then computed. Additionally, we evaluate \texttt{RadGraph Precision, Recall, and F1} using the RadGraph model ~\cite{jain2021radgraph}, which parses reports into graphs of clinical entities (anatomical references and observations) and their relations, followed by calculating the overlap between the entities and relations in the generated and reference reports.\\
\textbf{RadCliQ Metrics.} \texttt{RadCliQ} integrates \texttt{RadGraph F1} and \texttt{BLEU} scores to estimate the clinical error rate, providing a holistic quality assessment of the generated report. This metric closely aligns with radiologists' evaluations of report quality. For evaluation, we report both version 0 and version 1 of \texttt{RadCliQ}.\\
\textbf{GPT-based Evaluator.} The \texttt{GREEN} metric is an open-source evaluation framework for radiology report generation. It calculates matching findings and error counts between the generated report and the ground-truth report, providing clinical accuracy scores, while using a smaller language model for efficiency.

\subsection*{Sentence Factual Precision}
\begin{table*}[!htbp]
\centering
\scriptsize 
\renewcommand{\arraystretch}{1.2} 
\setlength{\tabcolsep}{3pt} 
\adjustbox{max width=\textwidth}{
\begin{tabular}{lccccccccc}
\toprule
\textbf{Sentence Removal} & \textbf{BLEU Score$\uparrow$} & \textbf{BERTScore$\uparrow$} & \textbf{Semb Score$\uparrow$} & \textbf{RadGraph Recall$\uparrow$} & \textbf{RadGraph Precision$\uparrow$} & \textbf{RadGraph Combined$\uparrow$} & \textbf{GREEN$\uparrow$} & \textbf{RadCliQ-v0$\downarrow$} & \textbf{RadCliQ-v1$\downarrow$} \\
\midrule
\textbf{Original Report}           & 0.1836 & 0.3995 & 0.4075 & 0.1801         & 0.2150 & 0.1874         & 0.2942 & 3.3838 & 1.1520 \\
\textbf{3\% Removed}      & 0.1814(0.1800)          & 0.3994(0.3915)          & 0.4082(0.4014)          & 0.1787(0.1751)          & 0.2166(0.2154)          & 0.1875(0.1842)          & 0.2936          & 3.3810(3.4217)          & 1.1513(1.1752)          \\
\textbf{5\% Removed }      & 0.1792(0.1763)          & 0.3990(0.3893)         & 0.4092(0.3982)          & 0.1778(0.1720)          & 0.2185(0.2154)          & \textbf{0.1878}(0.1822)          & 0.2929          & 3.3782(3.4348)          & \textbf{1.1506}(1.1845)          \\
\textbf{7\% Removed}      & 0.1758(0.1722)          & 0.3983(0.3869)          & 0.4091(0.3943)          & 0.1763(0.1681)         & 0.2202(0.2156)          & 0.1876(0.1795)          & 0.2924          & 3.3786(3.4508)          & 1.1522(1.1957)          \\
\textbf{9\% Removed }      & 0.1713(0.1693)          & 0.3971(0.3857)          & 0.4106(0.3942) & 0.1753(0.1655)          & 0.2237(0.2169)          & 0.1877(0.1782)         & 0.2904          & 3.3758(3.4545) & 1.1524(1.1994)          \\
\textbf{11\% Removed}      & 0.1678(0.1653)          & 0.3963(0.3827)          & 0.4116(0.3863)         & 0.1736(0.1611)          & 0.2263(0.2153)          & 0.1875(0.1749)          & 0.2889         & 3.3742(3.4805)          & 1.1530(1.2162)          \\
\textbf{13\% Removed }      & 0.1650(0.1615)          & 0.3959(0.3805)          & \textbf{0.4121}(0.3843)         & 0.1717(0.1584)          & \textbf{0.2287}(0.2165)          & 0.1871(0.1736)          & 0.2882          & \textbf{3.3729 }(3.4905)         & 1.1536(1.2236)          \\
\bottomrule
\end{tabular}
}
\caption{Comparison of various metrics across different levels of pruning guided by sentence uncertainty. Values in \textbf{bold} indicate metrics that improve the most compared to the original reports. An up arrow (\(\uparrow\)) signifies that a higher value is better for the metric, while a down arrow (\(\downarrow\)) indicates that a lower value is preferable. Values in parentheses show results from a baseline of random pruning at the same level.}
\label{tab:pruning_comparison}

\end{table*}
We use RadGraph to compute the sentence-level factual precision score for UQ evaluation. Following the notation from Section~\ref{sec:method}, we calculate the precision for each sentence $s_{ij}$, where $s_{ij}$ represents the $j$-th sentence in the generated report $\hat{r}_{i}$. 
\begin{eqnarray}
\label{eq:sentence_precision} 
P_\textbf{sentence}\left(s_{ij}, r_{i} \right) \nonumber &=& \frac{\left| g\left( s_{ij} \right) \cap g\left( r_i \right) \right|}
{\left| g\left( s_{ij} \right) \right|} \nonumber\\
&=& \frac{\left| \overline{V}_{ij} \cap \overline{V}_{i} \right|}
{\left| \overline{V}_{ij} \right|}
\end{eqnarray}
For edge cases in our experimented LLM model, such as the sentence "The patient is status post sternotomy," RadGraph fails to parse and returns an empty pair set. We flag these cases with a precision score of negative one \footnote{Correspondingly, we assign an uncertainty score of $u$=1 for these sentences during uncertainty calculation.}, indicating the lowest possible precision. This approach is reasonable, as such sentences typically refer to non-existent prior exams.

\subsection*{Sentence Abstention}
Given the positive results of report-level UQ in improving factual accuracy through report abstention, we extended this approach to sentence-level abstention to assess whether removing high-uncertainty sentences across the dataset could improve performance across various metrics. Table~\ref{tab:pruning_comparison} presents the results.

As the percentage of sentence abstention increases, we observe a drop in lexical scores such as BLEU and RadGraph recall, while GREEN remains consistent. Notably, RadGraph precision and RadCliQ metrics demonstrate improvement, indicating that selectively removing high-uncertainty sentences leads to higher factual precision in these aspects. Compared to the baseline of random sentence removal at the same levels, our uncertainty-guided abstention consistently yields superior results, indirectly demonstrating that our sentence-level UQ effectively identifies sentences with low factual accuracy.

However, unlike report-level abstention, sentence-level abstention may be less practical due to the complex nature of radiology reports, where sentences often contain multiple clinical claims. Removing entire sentences risks omitting relevant information, making this approach too coarse for practical application. In future work, we aim to integrate sentence-level UQ into the generation process itself, enabling more granular control to enhance factual accuracy without the need for sentence removal.

\end{document}